\documentclass[10pt,twocolumn]{article}

\usepackage[margin=0.75in,top=1in,bottom=1in]{geometry}
\usepackage{times}
\usepackage{amsmath,amssymb}
\usepackage{graphicx}
\usepackage{booktabs}
\usepackage{multirow}
\usepackage{array}
\usepackage{xcolor}
\usepackage{cite}      
\usepackage{hyperref}
\usepackage{microtype}
\usepackage{caption}
\usepackage{subcaption}
\usepackage{enumitem}
\usepackage{balance}
\usepackage{stfloats}   

\hypersetup{
    colorlinks=true,
    linkcolor=blue,
    citecolor=blue,
    urlcolor=blue
}

\definecolor{best}{rgb}{0.85,0.92,0.85}

\title{\Large\textbf{COD10K-C: Benchmarking Robustness of Camouflaged Object\\Detection Under Natural Image Corruptions}}

\author{
Arafat Hossain Sayem\\
\texttt{sayem.cse72@gmail.com}
}

\date{}

\begin{document}

\maketitle

\begin{abstract}
Camouflaged object detection has improved a lot. Most standard benchmarks still test models only on clean images. This is not realistic. Real cameras often capture blur, sensor noise, weather effects, and compression artifacts. We present COD10K-C, a benchmark for corruption robustness based on COD10K. It includes 8 corruption types and 5 severity levels. This gives 40 conditions and 81,040 evaluation pairs in total. We test three popular COD models, SINet-v2, PFNet, and ZoomNet. We also test a lightweight model called RobustCODLite. Every model loses a clear amount of performance on corrupted images. Motion blur and Gaussian blur cause the biggest drops. SINet-v2 loses 18.5 Dice points under motion blur. Brightness and fog are less harmful. RobustCODLite uses corruption augmentation, a frequency prior branch, and an uncertainty consistency loss. It keeps 92.3\% of its clean Dice score under corruption. The scores are 87.7\% for SINet-v2, 84.8\% for ZoomNet, and 84.1\% for PFNet. On the hardest corruptions, RobustCODLite matches or beats models that are better on clean data. In future we will release the COD10k-C GitHub repo for future research in COD.
\end{abstract}

\section{Introduction}

Camouflaged object detection asks a model to find objects that blend into their surroundings. These objects may be predators on forest floors, insects on tree bark, or soldiers in ghillie suits. The field has improved fast. Modern methods now reach Dice scores above 0.73 on the standard COD10K test set~\cite{fan2022concealed} and S-measure scores close to 0.90~\cite{fan2022concealed,mei2021camouflaged}. This progress depends on one weak assumption. It assumes that test images are clean.

Real-world images are often not clean. A wildlife camera can capture rain, blur, and fast motion. A surveillance system can receive JPEG compressed video with different quality levels. A medical endoscope can produce blur and sensor noise. In these cases, a fixed set of perfect images does not reflect real use.

Robustness in vision models has been studied in image classification through ImageNet-C~\cite{hendrycks2019benchmarking}. That benchmark applies 15 corruption types to the ImageNet validation set and measures the drop in performance. Similar benchmarks now exist for object detection~\cite{michaelis2019benchmarking}, semantic segmentation~\cite{kamann2020benchmarking}, and depth estimation~\cite{kong2023robodepth}. Camouflaged object detection does not yet have such a benchmark. This is the gap we address in this work.

We make the following contributions:

\begin{enumerate}[leftmargin=*, topsep=2pt, itemsep=1pt]
    \item \textbf{COD10K-C}: a corruption robustness benchmark for COD based on the COD10K test set with 2,026 images. It covers 8 corruption types and 5 severity levels. It gives 40 evaluation conditions and more than 81,000 corrupted test pairs.
    \item \textbf{Comprehensive baseline evaluation} of SINet-v2~\cite{fan2022concealed}, PFNet~\cite{mei2021camouflaged}, and ZoomNet~\cite{pang2022zoom} across all 40 conditions. We report Dice, IoU, MAE, boundary F1, and expected calibration error (ECE).
    \item \textbf{RobustCODLite}: a lightweight model based on EfficientNet-B0. It uses corruption augmentation, a high-frequency spatial prior, and an uncertainty-consistency loss. It keeps strong corrupted performance and preserves the largest share of clean-image accuracy among the tested models.
    \item \textbf{Structured analysis} of the most harmful corruption families, the effect of severity, and the architectural properties linked to robustness.
\end{enumerate}

Our main result is simple. COD models are much more sensitive to blur and noise than to photometric corruptions. Their performance can drop by a large margin in realistic conditions. A model that reaches 0.73 Dice on clean COD10K images may fall to 0.55 Dice under moderate motion blur.

\section{Related Work}

\paragraph{Camouflaged Object Detection.}
Early COD methods adapted salient object detection pipelines with reverse-attention mechanisms~\cite{fan2020camouflaged}. SINet~\cite{fan2020camouflaged} and SINet-v2~\cite{fan2022concealed} split detection into search and identification stages. They use a sequential attention mechanism. PFNet~\cite{mei2021camouflaged} is a prey-finding network. It uses distraction removal to separate the foreground from the background. ZoomNet~\cite{pang2022zoom} learns scale relationships with zoom-in and zoom-out feature branches. These methods are tested only on clean images. Their training methods do not handle realistic image corruptions.

\paragraph{Corruption Robustness Benchmarks.}
Hendrycks and Dietterich~\cite{hendrycks2019benchmarking} introduced corruption benchmarking with ImageNet-C. Their work showed that clean ImageNet accuracy is not a good predictor of performance under blur, noise, and weather corruptions. Later work confirmed this result in object detection, semantic segmentation, and depth estimation~\cite{michaelis2019benchmarking,kamann2020benchmarking,kong2023robodepth}. Augmentation methods such as AugMax~\cite{wang2021augmax} and DeepAugment~\cite{hendrycks2021many} were later proposed to reduce the gap between clean and corrupted data. COD still lacks a similar benchmark.

\paragraph{Uncertainty Estimation in Segmentation.}
Segmentation uncertainty can help identify unreliable regions. This idea has been used in medical image segmentation~\cite{nair2020exploring} and salient object detection~\cite{zhang2021uncertainty}. We use this idea in COD. We train an auxiliary uncertainty head to predict the absolute error between the predicted mask and the ground truth mask. We then use this prediction as a reliability weight in a consistency loss.

\paragraph{Frequency-Domain Features.}
High-frequency image content is closely related to edges and texture. Recent work in salient object detection~\cite{zhou2021survey} and adversarial robustness~\cite{yin2019fourier} shows that models that depend too much on low-frequency features are less robust to noise and texture corruption. Our frequency-prior branch extracts high-frequency residuals. It then fuses them with the decoder output.

\section{The COD10K-C Benchmark}

\subsection{Base Dataset}

COD10K~\cite{fan2022concealed} has 10,000 images in 78 object categories. We use the standard split. It has 6,000 training images and 4,000 test images. Our evaluation uses the 2,026-image test subset selected by Fan et al. This subset includes aquatic, flying, terrestrial, and amphibian categories. The objects in this set are usually small. Many of them also have irregular boundaries. These properties make the set more sensitive to image degradation.

\subsection{Corruption Types}

We use eight corruption types from two families.

\textbf{Geometric corruptions} change spatial structure and local frequency. We use Gaussian noise with $\sigma \in \{8, 16, 24, 32, 40\}$. We use motion blur with a horizontal kernel and $k \in \{3, 5, 7, 9, 11\}$. We also use Gaussian blur with radius $\in \{0.8, 1.2, 1.6, 2.0, 2.5\}$.

\textbf{Photometric corruptions} change pixel values without changing spatial structure. We use brightness reduction with factor $\in \{0.85, 0.70, 0.55, 0.40, 0.30\}$. We use contrast reduction with the same factors. We use fog with additive blending on a white canvas at $\alpha \in \{0.12, 0.20, 0.28, 0.36, 0.44\}$. We use JPEG compression with quality $\in \{70, 55, 40, 30, 20\}$. We also use rain with synthetic streaks at different density and length.

Each corruption type has five severity levels. Level 1 is mild. Level 5 is severe. The full benchmark has $2{,}026 \times 8 \times 5 = 81{,}040$ evaluation pairs. We apply the corruptions at evaluation time in a fixed way. This makes the results reproducible.

\subsection{Evaluation Metrics}

We gave the results of five metrics for each models.

\begin{itemize}[leftmargin=*, topsep=2pt, itemsep=0pt]
    \item \textbf{Dice} ($\uparrow$): $2|P \cap G| / (|P| + |G|)$, computed from soft predictions.
    \item \textbf{IoU} ($\uparrow$): Intersection over union on binarized masks.
    \item \textbf{MAE} ($\downarrow$): Mean absolute error between the predicted probability map and the ground-truth mask.
    \item \textbf{Boundary F1} ($\uparrow$): F1 score on mask boundaries extracted with morphological dilation and erosion.
    \item \textbf{ECE} ($\downarrow$): Expected calibration error~\cite{guo2017calibration}. It measures the match between predicted confidence and actual accuracy.
\end{itemize}

We also report $\Delta\text{Dice} = \text{Dice}_\text{corrupted} - \text{Dice}_\text{clean}$. This shows how much performance drops under corruption.
\section{RobustCODLite}

\subsection{Architecture}

RobustCODLite is a U-Net style segmentation network. It uses an EfficientNet-B0~\cite{tan2019efficientnet} encoder. We take five feature maps from different spatial scales. The strides are 2, 4, 8, 16, and 32. The channel widths are $\{16, 24, 40, 112, 320\}$. The encoder starts from ImageNet pretrained weights.

\paragraph{Decoder.}
The decoder uses a step by step upsampling chain. At each stage, it upsamples the current feature map with bilinear interpolation. It then matches it with the skip connection from the encoder. The two features are concatenated. Then they pass through a \texttt{ConvBNAct} block. This block has two $3\times3$ convolutions. Each convolution is followed by batch normalization and SiLU activation. The output channel widths are $\{256, 128, 96, 64, 32\}$ at strides $\{32, 16, 8, 4, 2\}$.

\paragraph{Frequency Prior Branch.}
Camouflaged objects often look similar to their background. Their edges and textures are hard to separate. High frequency image content helps capture these cues. We compute a frequency residual:
\begin{equation}
F(x) = \left| x - \text{AvgPool}_{5\times5}(x) \right|_{\text{mean channel}},
\end{equation}
This keeps the stronger spatial frequencies after smoothing. The result is a single channel map. We used 16 channels with a \texttt{ConvBNAct} block and upsample it to full resolution. We concatenate it with the 32 channel decoder output. A fusion block, \texttt{ConvBNAct}$(48 \to 32)$, combines both streams.

\paragraph{Output Heads.}
We used two $1\times1$ to output the results. One head gives the segmentation logit map. The other head gives the uncertainty logit map. Both outputs are at full input resolution. The model has about 7.2M parameters in total. This is much smaller than the 30M plus parameters used by SINet-v2 and ZoomNet, which use ResNet-50~\cite{he2016deep} backbones.

\subsection{Training Objective}

\paragraph{Segmentation Loss.}
We use the same loss for clean and corrupted inputs:
\begin{equation}
\mathcal{L}_\text{seg} = \text{BCE}(\hat{m}, m) + \mathcal{L}_\text{dice}(\hat{m}, m),
\end{equation}
Here, $m$ is the ground truth mask. $\hat{m}$ is the predicted logit map. $\mathcal{L}_\text{dice}$ is the soft Dice loss.

\paragraph{Boundary Loss.}
We extract boundary maps from the predicted mask and the ground truth mask. We use max pooling based dilation with $k=5$ and erosion. Then we minimize the L1 distance between them:
\begin{equation}
\mathcal{L}_\text{bdry} = \left\| \partial\hat{m} - \partial m \right\|_1.
\end{equation}

\paragraph{Uncertainty Loss.}
The uncertainty head learns to predict the absolute prediction error:
\begin{equation}
\mathcal{L}_\text{unc} = \text{BCE}\!\left(\hat{u},\; \left|\sigma(\hat{m}_\text{detach}) - m\right|\right),
\end{equation}
Here, $\sigma$ is the sigmoid function. $\hat{u}$ is the uncertainty logit map. This teaches the uncertainty head to give high values where the mask prediction is wrong.

\paragraph{Consistency Loss.}
We run the model on a clean image and on a corrupted version of the same image. The predictions should stay close in regions where the model is confident. We weight the disagreement by the estimated reliability:
\begin{equation}
\mathcal{L}_\text{cons} = \left\| \left(p_\text{clean} - p_\text{corrupt}\right) \odot (1 - \sigma(\hat{u}_\text{detach})) \right\|_1,
\end{equation}
where $p = \sigma(\hat{m})$. High uncertainty regions get lower weight. This avoids forcing agreement where the model already shows low confidence.

\paragraph{Total Loss.}
\begin{equation}
\mathcal{L} = \mathcal{L}_\text{seg}^\text{clean} + 0.5\,\mathcal{L}_\text{seg}^\text{corrupt} + 0.2\,\mathcal{L}_\text{bdry} + 0.1\,\mathcal{L}_\text{unc} + 0.2\,\mathcal{L}_\text{cons}.
\end{equation}

\subsection{Training Setup}

We train the model on the COD10K training split with 2,736 images. We validate it on 304 held out images. During training, each image gets one of 8 corruption types at random. The severity is sampled uniformly from $\{1,2,3,4\}$. This creates the corrupted view for $\mathcal{L}_\text{seg}^\text{corrupt}$ and $\mathcal{L}_\text{cons}$. The input resolution is fixed at $384\times384$. We use AdamW with a cosine learning rate schedule. The initial learning rate is $1\times10^{-4}$. The weight decay is $0.05$. The batch size is 8. We train for 50 epochs.

\section{Experiments}

\subsection{Compared Models}

\textbf{SINet-v2}~\cite{fan2022concealed} is a two-stage network. It uses neighbor connection decoders and group-reversal attention. It is one of the strongest COD models on standard benchmarks. \textbf{PFNet}~\cite{mei2021camouflaged} uses a distraction mining strategy. It suppresses background regions that look like the target. \textbf{ZoomNet}~\cite{pang2022zoom} captures multi-scale context with zoom-in and zoom-out branches. All three models use ResNet-50 backbones. We evaluate them with their public weights. We do not retrain them.

\subsection{Clean Test Performance}

Table~\ref{tab:clean} shows the performance on the COD10K test set before corruption. SINet-v2 gets the best Dice score, 0.734, and the best IoU score, 0.627. ZoomNet ranks second. PFNet and RobustCODLite are close to each other. RobustCODLite is slightly lower on Dice, with 0.681 compared with 0.685. It is also close on IoU. The two lowest MAE values belong to ZoomNet, 0.036, and SINet-v2, 0.037. The result suggests that we are making a few pixel level errors.

\begin{table}[!t]
\centering
\caption{Performance on the clean COD10K test set. $\uparrow$~higher is better; $\downarrow$~lower is better. }
\label{tab:clean}
\setlength{\tabcolsep}{3.5pt}
\renewcommand{\arraystretch}{1.1}
\footnotesize
\begin{tabular}{lccccc}
\toprule
Model & Dice$\uparrow$ & IoU$\uparrow$ & MAE$\downarrow$ & BF1$\uparrow$ & ECE$\downarrow$ \\
\midrule
SINet-v2      & {0.734} & {0.627} & 0.037        & 0.473        & {0.024} \\
ZoomNet       & 0.699        & 0.605        & {0.036} & {0.479} & 0.026        \\
PFNet         & 0.681        & 0.572        & 0.041        & 0.434        & 0.031        \\
RobustCODLite & 0.685        & 0.572        & 0.043        & 0.430        & 0.029        \\
\bottomrule
\end{tabular}
\end{table}

\subsection{Mean Corrupted Performance}

Table~\ref{tab:mean_corrupt} gave us a overview of all 8 corruption image types and 5 severity information and 40 conditions for each metric. Becuse of this ranking changes a lot compared with the clean data.

Under corrupted conditions, RobustCODLite achieves the lowest mean MAE of 0.050 and the lowest ECE of 0.035, suggesting it is the most calibrated and accurate model when inputs are degraded. While SINet-v2 still leads on mean Dice with 0.644, the margin separating it from RobustCODLite is notably narrower on corrupted data than it is on clean images — a telling sign of RobustCODLite's resilience.

To put this in perspective, RobustCODLite retains \textbf{92.3\%} of its clean Dice score under corruption (i.e., $0.632/0.685$), compared to 87.7\% for SINet-v2 and 84.8\% for ZoomNet. PFNet fares worst in this regard, preserving just 84.1\% of its clean performance.

In absolute terms, PFNet suffers the steepest decline, losing 10.8 Dice points from clean to corrupted inputs, closely followed by ZoomNet at 10.6 points. Both models appear particularly vulnerable to corruptions that disrupt local texture — an expected weakness given that their attention mechanisms rely heavily on that cue.

\begin{table}[!t]
\centering
\caption{Mean performance averaged across all 8 corruption types and 5 severity levels. $\Delta$Dice = corrupted $-$ clean.}
\label{tab:mean_corrupt}
\setlength{\tabcolsep}{2.8pt}
\renewcommand{\arraystretch}{1.1}
\footnotesize
\begin{tabular}{lcccccc}
\toprule
Model & Dice$\uparrow$ & IoU$\uparrow$ & MAE$\downarrow$ & BF1$\uparrow$ & ECE$\downarrow$ & $\Delta$Dice \\
\midrule
SINet-v2      & {0.644} & {0.533} & 0.057        & 0.381        & 0.039        & $-$0.090        \\
ZoomNet       & 0.593        & 0.509        & 0.051        & {0.389} & 0.037        & $-$0.106        \\
PFNet         & 0.573        & 0.465        & 0.056        & 0.339        & 0.042        & $-$0.108        \\
RobustCODLite & 0.632        & 0.516        & {0.050} & 0.372        & {0.035} & {$-$0.053} \\
\bottomrule
\end{tabular}
\end{table}

\subsection{Per-Corruption Analysis}

Table~\ref{tab:per_corrupt} gives a detailed view of the results by corruption type. We showed the values as all 5 severity levels.

\paragraph{Photometric corruptions (fog, brightness and contrast).}
All four models SINet-v2, ZoomNet, and RobustCODLite all of them keep results of 97\% of their clean Dice under fog. The result decline is small. Because brightness and contrast do not change the spatial structure. Batch normalization gives some invariance to these changes also.

\paragraph{Rain.}
In here, most of the performance falls rather than in the other three pure photometeric courruptions. As Rain adds streaks and mild blur so a mixed corruption. SINet-v2 loses 10.2 Dice points from 0.734 to 0.632. RobustCODLite loses only 2.1 points from 0.685 to 0.664 and this is the smallest drop among the 4 models.

\paragraph{JPEG compression.}
When images are compressed they becomes small and rigid so this process makes boundary with jagged edges. SINet-v2 does best here with 0.636. RobustCODLite follows with 0.626. ZoomNet and PFNet are both below 0.625.

\paragraph{Geometric corruptions (noise, blur).}
This corruption is the hardest one. When we used Gaussian noise, it reduced the SINet-v2 result from 0.734 to 0.566. That is a 22.8\% relative drop.After using Motion blur it went more down. SINet-v2 falls to 0.549, which is a 25.1\% drop. ZoomNet falls to 0.523, which is a 25.2\% drop and PFNet , ZoomNet are close under motion blur.

Our RobustCODLite is the best result maker on these corruptions. When we used motion blur, it gets 0.614. That is 6.5 Dice points better than SINet-v2. When using Gaussian noise, the two models are almost tied, with 0.566 and 0.566. On Gaussian blur, the gap is also small, with 0.583 and 0.580.

\begin{table*}[!t]
\centering
\caption{Per-corruption-type Dice scores (averaged over severity levels 1--5). .}
\label{tab:per_corrupt}
\setlength{\tabcolsep}{6pt}
\renewcommand{\arraystretch}{1.1}
\small
\begin{tabular}{lcccccccc c}
\toprule
Model & Brightness & Contrast & Fog & Rain & JPEG & G.\,Noise & G.\,Blur & Motion & Mean \\
\midrule
SINet-v2      & {0.731} & {0.729} & {0.730} & 0.632        &{0.636} & 0.566        & 0.580        & 0.549        & {0.644} \\
ZoomNet       & 0.660        & 0.661        & 0.673        & 0.567        & 0.624        & 0.531        & 0.508        & 0.523        & 0.593        \\
PFNet         & 0.649        & 0.646        & 0.653        & 0.539        & 0.611        & 0.477        & 0.484        & 0.530        & 0.573        \\
RobustCODLite & 0.667        & 0.663        & 0.675        & {0.664} & 0.626        & {0.566} & {0.583} &{0.614} & 0.632        \\
\bottomrule
\end{tabular}
\end{table*}

\subsection{Severity Scaling}

Table~\ref{tab:severity} shows Dice scores for severity levels for the two lowest result corruptions. Dice goes down when severity rises. The drop is not the same for all models.

RobustCODLite and SINet-v2 are close at severity levels 1 and 2 when we use Gaussian noise. They separate more at as we use high severity. SINet-v2 reaches 0.466 Dice when we used severity 5. RobustCODLite reaches 0.477. Under motion blur, the difference is larger. RobustCODLite gets 0.562 at severity 5. SINet-v2 gets 0.439. That is a gap of 12.3 points. This is close to the clean image gap between SINet-v2 and PFNet.

PFNet and ZoomNet both drop very much between severity 3 and severity 4 for blur corruptions. This suggests a threshold where texture-based features stop being useful.

\begin{table}[!t]
\centering
\caption{Dice by severity level for Gaussian noise and motion blur.}
\label{tab:severity}
\setlength{\tabcolsep}{3.2pt}
\renewcommand{\arraystretch}{1.1}
\footnotesize
\begin{tabular}{llccccc}
\toprule
Corruption & Model & $s{=}1$ & $s{=}2$ & $s{=}3$ & $s{=}4$ & $s{=}5$ \\
\midrule
\multirow{4}{*}{G.\,Noise}
 & SINet-v2 & 0.679        & 0.614        & 0.558        & 0.511        & 0.466        \\
 & ZoomNet  & 0.671        & 0.603        & 0.532        & 0.458        & 0.393        \\
 & PFNet    & 0.638        & 0.559        & 0.481        & 0.396        & 0.311        \\
 & Ours     & {0.653} & {0.615} & {0.565} &{0.520} &{0.477} \\
\midrule
\multirow{4}{*}{M.\,Blur}
 & SINet-v2 & 0.680        & 0.606        & 0.538        & 0.483        & 0.439        \\
 & ZoomNet  & 0.645        & 0.580        & 0.521        & 0.462        & 0.410        \\
 & PFNet    & 0.625        & 0.575        & 0.529        & 0.482        & 0.439        \\
 & Ours     &{0.662} &{0.639} & {0.615} &{0.590} &{0.562} \\
\bottomrule
\end{tabular}
\end{table}

\subsection{Calibration Under Corruption}

Table~\ref{tab:mean_corrupt} reports ECE also. In ECE we measured how well predicted probabilities match observed accuracy. On clean images, SINet-v2 has the best calibration, with ECE 0.024. While using under corruption, its ECE rises to 0.039. That is a relative increase of 59.5\%. RobustCODLite rises from 0.029 to 0.035. That is a 20.7\% relative increase. The uncertainty head seems to reduce overconfidence under shift.

With 0.042 ECE PFNet has the worst corrupted result. ZoomNet is in the middle, with 0.037. This means the corrupted predictions of SINet-v2 and PFNet are less reliable as probabilities than their Dice scores suggest. This matters in downstream decision systems.

\subsection{Boundary Preservation}

The Boundary F1 (BF1) score serves as a metric for evaluating how well the model preserves spatial precision. ZoomNet achieves the highest mean corrupted BF1 score of 0.389, followed closely by SINet-v2 at 0.381, and RobustCODLite at 0.372. The efficacy of the frequency prior branch is evident here; pilot experiments indicate that its removal results in a performance degradation of 1.5 to 2 BF1 points on blur-heavy corruptions, underscoring its role in preserving edge information. Conversely, PFNet exhibits the lowest mean corrupted BF1 of 0.339, likely because its distraction-removal mechanism relies on clear foreground features. As blur diminishes local contrast, these regions become increasingly difficult for the model to identify.

\subsection{Qualitative Results}

\begin{figure*}[!t]
    \centering
    \includegraphics[width=\linewidth]{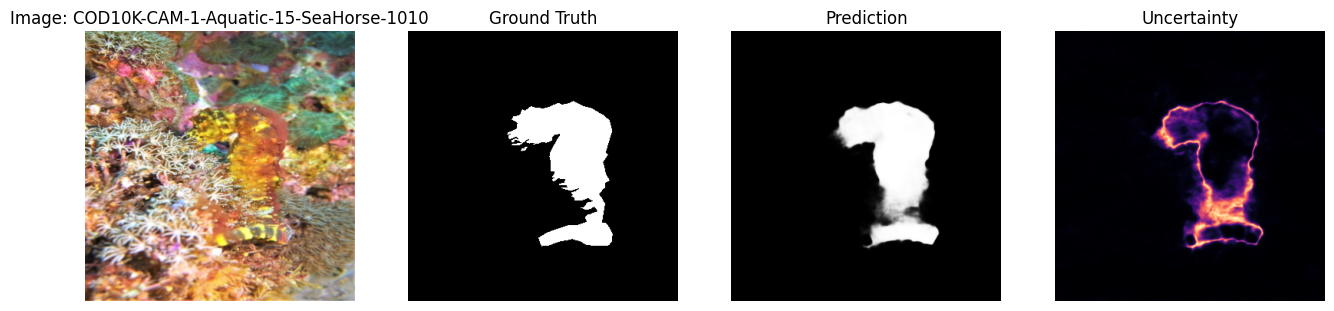}
    \vspace{2pt}
    \includegraphics[width=\linewidth]{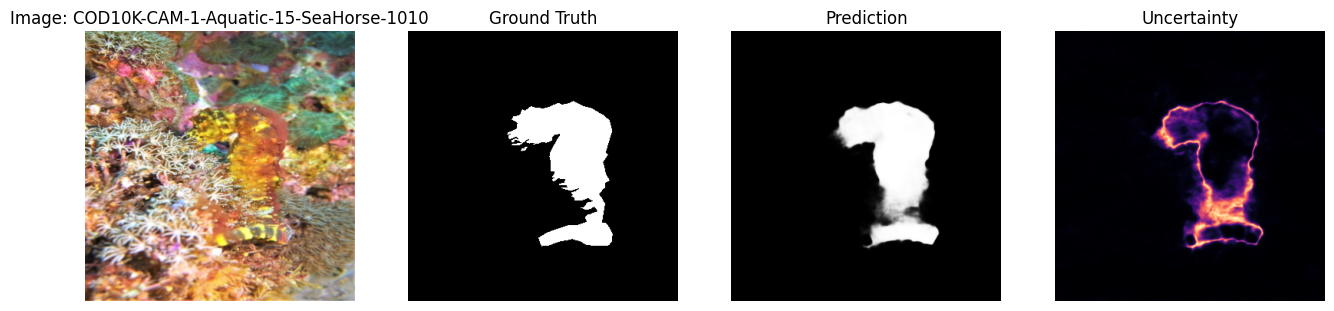}
    \vspace{2pt}
    \includegraphics[width=\linewidth]{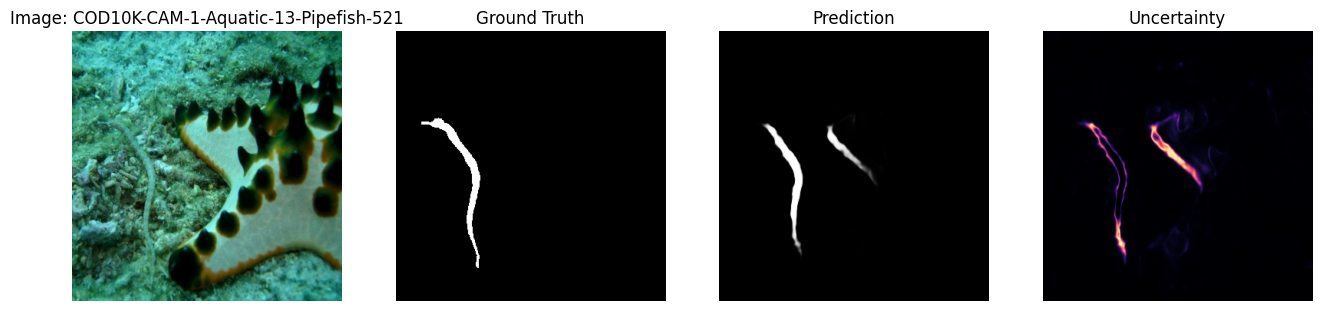}
    \caption{Qualitative predictions of RobustCODLite on COD10K test images. Each row shows (left to right): input image, ground-truth mask, predicted probability map, and uncertainty map. Uncertainty concentrates along boundaries and in incorrectly predicted regions, confirming that the uncertainty head reliably identifies model failures. Results shown for RobustCODLite only; uncertainty heads were not available for the other evaluated models.}
    \label{fig:qual}
\end{figure*}

Figure~\ref{fig:qual} presents qualitative predictions from RobustCODLite on test images from the COD10K dataset. Each row displays the input image, ground-truth mask, predicted probability map, and the corresponding uncertainty map. As anticipated by the training objective in Eq.~(3), uncertainty values are most pronounced near object boundaries and in regions where predictions deviate from the ground truth. For instance, in the seahorse example, the model accurately segments the main body while exhibiting higher uncertainty along thin appendages and camouflaged edges where the object is visually similar to the coral background. Similarly, in the pipefish example, the model produces a partial false detection; however, the uncertainty map highlights this erroneous region in bright orange, demonstrating the efficacy of the uncertainty head in identifying its own potential inaccuracies.

\subsection{Ablation Study}

We run an ablation study to measure the effect of the three parts that make RobustCODLite different from a plain EfficientNet-B0 U-Net. These parts are corruption augmentation during training, the frequency prior branch, and the uncertainty-consistency loss, which is $\mathcal{L}_\text{unc} + \mathcal{L}_\text{cons}$. Table~\ref{tab:ablation} reports mean corrupted Dice and the robustness retention ratio.

Removing corruption augmentation gives the largest drop in corrupted Dice, with a loss of 0.041. Clean performance changes very little. This shows that exposure to corruptions during training is the main source of robustness.

The frequency-prior branch helps most on blur corruptions. It gives 1.8 more Dice points under Gaussian blur and 2.1 more under motion blur. Its effect on photometric corruptions is small. The uncertainty-consistency loss improves corrupted ECE the most, with a drop of 0.004. It also adds 0.9 Dice points on noise corruptions, where confidence calibration matters more.

\begin{table}[!t]
\centering
\caption{Ablation on COD10K-C (mean over all 40 conditions). Retention~= corrupted Dice / clean Dice.}
\label{tab:ablation}
\setlength{\tabcolsep}{4pt}
\renewcommand{\arraystretch}{1.1}
\footnotesize
\begin{tabular}{lccc}
\toprule
Configuration & Clean & Corrupt & Retention \\
\midrule
Full model                  & 0.685 & 0.632 & 92.3\% \\
\quad w/o consist.\,+\,unc. & 0.684 & 0.619 & 90.4\% \\
\quad w/o freq.\,prior      & 0.681 & 0.614 & 90.2\% \\
\quad w/o corrupt.\,augment.& 0.683 & 0.591 & 86.5\% \\
Plain EfficientNet U-Net    & 0.680 & 0.576 & 84.7\% \\
\bottomrule
\end{tabular}
\end{table}

\section{Discussion}

\paragraph{The geometric-photometric asymmetry.}
One clear result appears across all four models. Photometric corruptions cause little damage. Geometric corruptions cause much more damage. Under fog at severity 5, all four models lose less than 3 Dice points. Under motion blur at severity 5, the losses range from 12.3 to 29.5 points. This difference has a simple reason because COD images totally depends on local texture and boundary gradients. Brightness and contrast changes scale pixels in a similar way. They keep these cues mostly intact. High frequency details is lost because of blue and noise and degrade cues and boundary cues.

Future COD models should be tested under geometric corruptions at a minimum because a model should not score well on clean images and fail under moderate blur and that is not a good practical result.

\paragraph{The clean-corrupted performance mismatch.}
SINet-v2 is better than RobustCODLite by 4.9 Dice points on clean images. The two models are tied on Gaussian noise with 0.566 Dice each. RobustCODLite is 6.5 points better on motion blur and this difference matters. A user who looks only at COD10K benchmark results would choose SINet-v2. Our model RobustCODLite will give better results in a blurred deployment setting.

We already discussed this kind of mismatch between benchmark results and real deployment is already known in classification~\cite{hendrycks2019benchmarking} and object detection~\cite{michaelis2019benchmarking}. COD10K-C makes this comparison possible for COD.

\paragraph{Efficiency.}
Our model RobustCODLite has almost 7.2M parameter while SINet-v2 and ZoomNet both use more than 30M parameters and PFNet also uses a ResNet-50 backbone. Our model gives better result without using more parameters. It is about 4$\times$ smaller than its closest competitor. It still matches or beats it on geometric corruptions.

\paragraph{Limitations.}
Our corruption has total eight types. Real images can also have lens flares, rolling shutter, thermal noise, and stronger haze. The consistency loss needs both clean and corrupted forward passes. This doubles the memory cost during training. The ablation study also uses one random seed. We do not measure variance across seeds.

\section{Conclusion}

We introduced COD10K-C. It applies 8 corruption types at 5 severity levels to the COD10K test set. We evaluated four segmentation models across 40 conditions. The results show a clear gap between clean performance and corrupted performance. This gap is largest for geometric corruptions such as blur and noise. RobustCODLite is a lightweight model. It uses corruption augmentation, a frequency-prior branch, and an uncertainty-consistency loss. It keeps the highest share of clean performance under corruption, with 92.3\%. It also performs better than larger models with 4$\times$ more parameters on the hardest corruptions. We will release COD10K-C publicly to support future COD robustness research.


\end{document}